\documentclass[letterpaper, 10 pt, conference]{ieeeconf}  

\IEEEoverridecommandlockouts                              

\usepackage{graphics} 
\usepackage{amsmath} 
\usepackage{amssymb}  
\usepackage{mathtools}
\usepackage{graphicx}
\usepackage{multirow}
\usepackage{algorithm,algpseudocode}
\usepackage[hidelinks]{hyperref}
\usepackage{array}
\usepackage{textcomp}
\usepackage{booktabs}
\usepackage[table,xcdraw]{xcolor}
\renewcommand{\arraystretch}{2}
\definecolor{lightgray}{gray}{0.95}

\title{\LARGE \bf
CLUE: Crossmodal disambiguation via Language-vision Understanding with attEntion
}

\author{%
Mouad Abrini and Mohamed Chetouani%
\\
\textit{Institut des Syst\`emes Intelligents et de Robotique (ISIR), Sorbonne Universit\'e} \\
Paris, France \\
{\ttfamily \{mouad.abrini, mohamed.chetouani\}@isir.upmc.fr}
}

\begin{document}
\bstctlcite{IEEEexample:BSTcontrol}

\maketitle
\thispagestyle{empty}
\pagestyle{empty}

\begin{abstract}
With the increasing integration of robots into daily life, human-robot interaction has become more complex and multifaceted. A critical component of this interaction is Interactive Visual Grounding (IVG), through which robots must interpret human intentions and resolve ambiguity.
Existing IVG models generally lack a mechanism to determine when to ask clarification questions, as they implicitly rely on their learned representations. CLUE addresses this gap by converting the VLM’s cross-modal attention into an explicit, spatially grounded signal for deciding when to ask. We extract text to image attention maps and pass them to a lightweight CNN to detect referential ambiguity, while a LoRA fine-tuned decoder conducts the dialog and emits grounding location tokens. We train on a real-world interactive dataset for IVG, and a mixed ambiguity set for the detector. With InViG-only supervision, our model surpasses a state-of-the-art method while using parameter-efficient fine-tuning. Similarly, the ambiguity detector outperforms prior baselines. Overall, CLUE turns the internal cross-modal attention of a VLM into an explicit, spatially grounded signal for deciding when to ask. The data and code are publicly available at: 
\textcolor{blue}{\href{mouadabrini.github.io/clue}{mouadabrini.github.io/clue}}
\end{abstract}

\section{INTRODUCTION}
\begin{figure}
    \centering
    \includegraphics[width=0.9\linewidth]{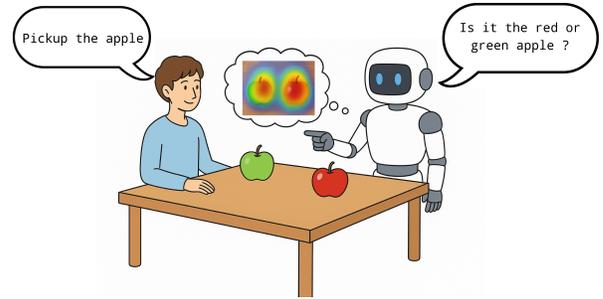}
    \caption{Problem illustration: when an instruction is underspecified, the robot should detect it and ask for clarification (AI generated, then edited)}
    \label{fig:intro}
\end{figure}

Robots operating in human spaces must interpret underspecified language instructions and act safely. A central capability is interactive visual grounding (IVG): given a natural-language instruction (e.g., “Pick up the apple”), the robot must identify the intended referent and, when necessary, ask for clarification (see Fig \ref{fig:intro}). While recent IVG systems can localize objects from referring expressions \cite{yu2018mattnetmodularattentionnetwork,kamath2021mdetrmodulateddetection,Jin_2023_CVPR,han2024zeroshotreferringexpressioncomprehension} and even engage in dialog \cite{shridhar_interactive_2018,DBLP:journals/corr/abs-1710-06280,mi_interactive_2020,thomason_improving_2019}, an open question is how to detect that the current instruction is ambiguous in the specific visual scene.
Classical referring expression comprehension (REC) methods, from modular attention networks to modern transformer detectors, largely assume uniqueness, optimizing for a single best match \cite{yu2018mattnetmodularattentionnetwork,kamath2021mdetrmodulateddetection}. Interactive approaches in HRI typically decide to ask using heuristics (e.g., multiple plausible candidates) or token-level uncertainty (e.g., entropy/confidence from a language or policy model), as seen in planners that query based on grounding confidence \cite{yang_interactive_2022}, and LLM-driven ambiguity classifiers like CLARA \cite{park2024claraclassifyingdisambiguatinguser}. Recent dialog-trained systems further learn when to ask from demonstration, e.g., ClearVQA \cite{jian-etal-2025-teaching} and uncertainty-timed clarification \cite{testoni2024askingrightquestionright}. These signals can be effective, but are policy-level (answer vs. ask) and only indirectly grounded in the spatial structure of the scene: they rarely indicate where confusion arises or which distractors compete for attention.
We argue that ambiguity in grounded instructions arises in the joint image-language representation, where textual tokens align with multiple plausible visual candidates. In such cross-modal spaces, the decoder's text queries attend to visual patches that encode scene structure and attributes. When several objects fit the instruction, the alignment becomes non-selective, distributing mass across competing regions. We exploit this property by extracting the model's cross-modal self-attention weights and training a convolutional network on these maps to classify ambiguity. This turns a latent alignment pattern into a spatial signal that both (i) indicates when the instruction is underspecified and (ii) localizes where the confusion lies for better interpretability.

\subsection{Contributions}
\begin{itemize}
    \item Spatial ambiguity detection: a CNN over cross-modal attention maps that explicitly detects and localizes referential ambiguity.
    \item A synthetic dataset generated using the Isaac Sim Simulator \cite{NVIDIA_Isaac_Sim} for multimodal ambiguity detection.
    \item End-to-end disambiguation (IVG): a fine-tuned VLM on InViG only \cite{zhang2023invigbenchmarkinginteractivevisual} (real-world dataset) yielding state-of-the-art IVG performance, outperforming TiO \cite{xu2024unifiedinteractivevisualgrounding} trained on InViG-only.
\end{itemize}
\begin{figure*}[t]
    \centering
    \includegraphics[width=0.8\linewidth]{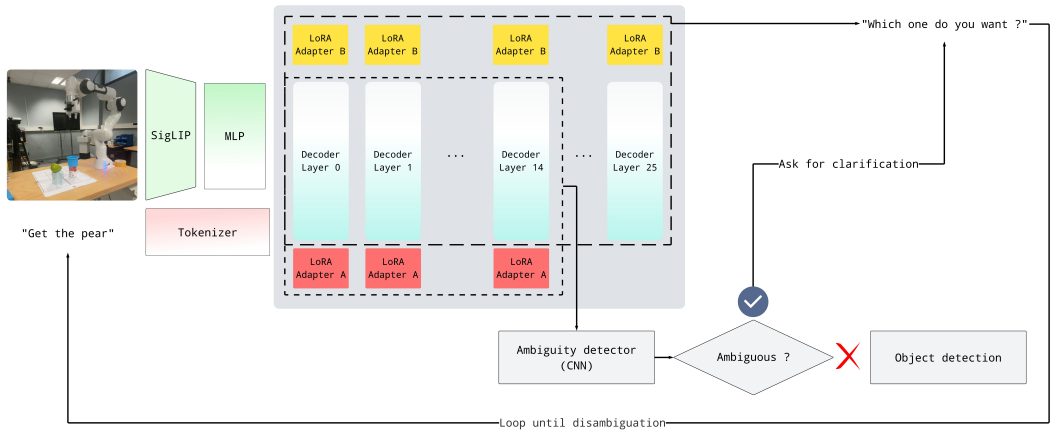}
    \caption{Overall CLUE architecture. An RGB image is encoded by SigLIP and projected by an MLP. The text prefix is tokenized and passed with the image tokens into a Gemma2 decoder equipped with LoRA adapters. The decoder both (i) autoregressively generates clarification questions and (ii) exposes cross-modal attention maps from mid-layers to a lightweight CNN ambiguity detect (Fig. \ref{fig:detectarch}). If the detector predicts ambiguous, the model asks a follow-up and updates the dialog context; otherwise it emits location tokens and triggers object detection. The loop repeats until disambiguation.}
    \label{fig:generalarch}
\end{figure*}
\section{RELATED WORK}
\subsection{Interactive Visual Grounding and Clarification Dialogues}
Ambiguity in referring expressions has long been addressed by interactive systems that can detect unclear references and ask users for clarification. Early human-robot interaction work integrated dialog or gestures to resolve ambiguity. For example, \cite{shridhar_interactive_2018} combined user pointing gestures with a dialog system to disambiguate spoken commands, and \cite{DBLP:journals/corr/abs-1710-06280} used a referring expression model to find candidate objects and then engaged in conversation to identify the correct target
\cite{mi_interactive_2020}. Similarly, \cite{ahn_interactive_2018} proposed generating spatial heatmaps and follow-up questions to locate the intended object, and \cite{thomason_improving_2019} showed that clarification questions from a robot improved the grounding of object references. In essence, if a user says “Bring me that cup,” an interactive agent must recognize the referential ambiguity (e.g. multiple cups are present) and ask a targeted question to clarify which cup is meant. Modern systems explicitly predict such ambiguity and generate clarification questions. For instance, \cite{yang_interactive_2022} presents an attribute-guided POMDP planner that uses vision-language grounding confidence to decide when to ask a question to differentiate between look-alike objects. Recent vision-language models are explicitly trained to ask when a question is ambiguous: \cite{jian-etal-2025-teaching} introduces the ClearVQA benchmark and methods that detect ambiguity, generate targeted clarification questions, and then answer after interaction. Likewise, \cite{testoni2024askingrightquestionright} analyzed human dialogs and model uncertainty to determine the “right time” to ask for clarification, aligning the agent’s questions with moments of referential uncertainty. \cite{zhang2023invigbenchmarkinginteractivevisual} introduce InViG, a large-scale benchmark with $\sim$ 500K human-robot interactions specifically targeting ambiguity in open-ended scenes. It evaluates how an agent should respond with clarifying questions in a natural manner rather than relying on fixed templates. 
\begin{figure*}[t]
    \centering
    \includegraphics[width=0.8\linewidth]{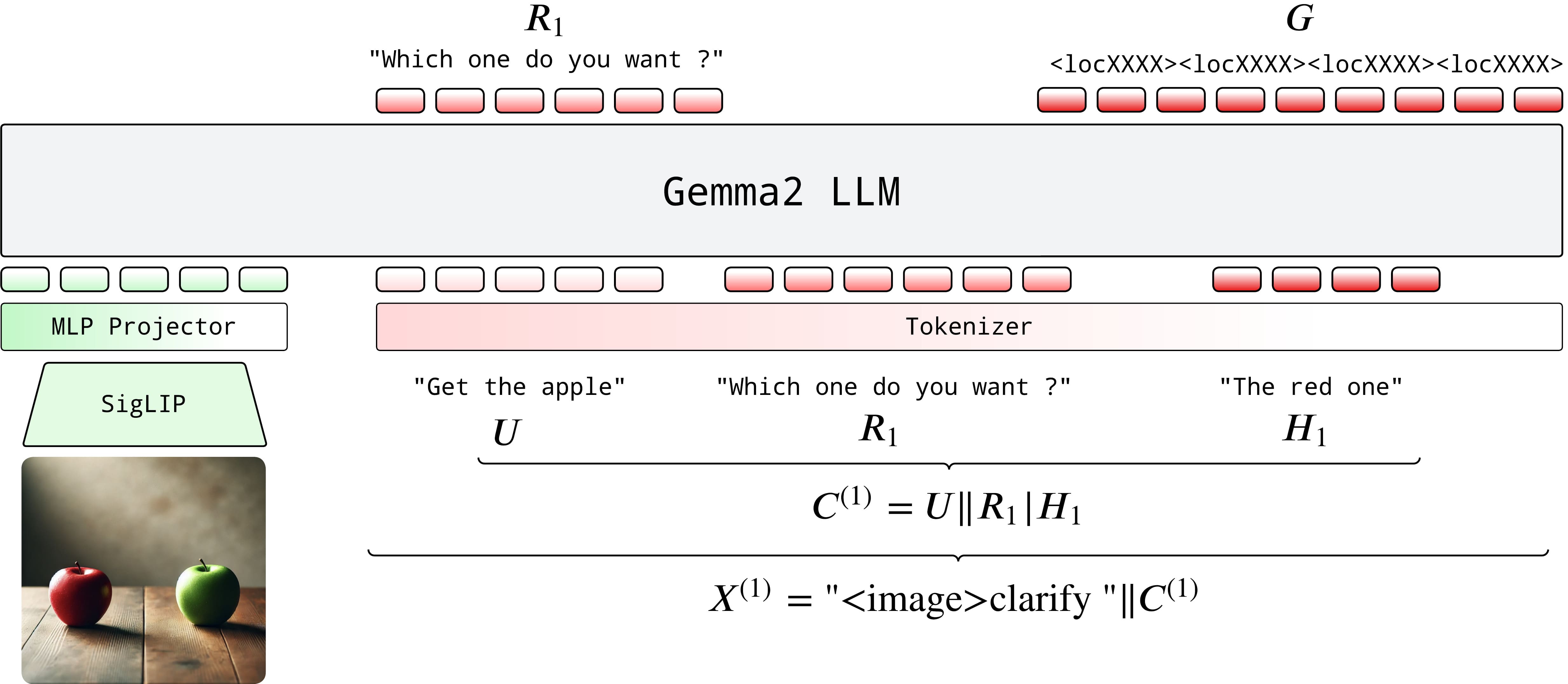}
    \caption{IVG with notation. An RGB image (AI generated in this example) is encoded by SigLIP and projected to the decoder via an MLP, the text prefix is tokenized and concatenated with image tokens. The Gemma2 LLM takes $X^{(1)}=''<image>\text{clarify} ''\Vert C^{(1)}$, where the running context after the first turn is $C^{(1)}=U\Vert R_1\Vert H_1$ (initial user request $U$, assistant question $R_1$, human reply $H_1$). The model autoregressively generates either a clarification segment (left stream, labeled $R_1$) or location tokens $G=<\text{locXXXX}>...$ (right stream) that decode to a bounding box. Green blocks denote image features, pink blocks denote text tokens.}
    \label{fig:ivgarch}
\end{figure*}
\cite{xu2024unifiedinteractivevisualgrounding} builds on that line by proposing TiO, an end-to-end model for interactive visual grounding in the wild and learns to gather information actively during the interaction. The main limitation of these kinds of methods is that they rely heavily on the statistical presence of the dialog rounds. For example, the InViG dataset contains at least one dialog round, so the model will always ask a clarification question, no matter what, even if the initial instruction is unambiguous.
All these approaches typically rely on model confidence scores or predicted likelihoods to trigger clarifications. In contrast, our method enables triggering clarifications using a spatial, cross-modal attention signal from a VLM, rather than token likelihoods or confidence heuristics.

\subsection{Ambiguity Detection in Grounded Instructions}
To detect ambiguity, models typically trigger clarification via heuristics (e.g., multiple candidate objects) or learned uncertainty (token/policy entropy, confidence). For example, the model proposed in \cite{park2024claraclassifyingdisambiguatinguser} classifies commands as clear/ambiguous/infeasible using an LLM-based confidence signal and asks for clarification when needed. Interactive grounding also uses dialog/gestures or heatmap/question strategies to resolve reference ambiguity \cite{shridhar_interactive_2018, DBLP:journals/corr/abs-1710-06280, mi_interactive_2020, ahn_interactive_2018, thomason_improving_2019}, and recent work learns when to ask from data; for instance, by planning over grounding confidence \cite{yang_interactive_2022} or using model uncertainty to guide questioning \cite{testoni2024askingrightquestionright}. To evaluate such capabilities, benchmarks like ClearVQA \cite{jian-etal-2025-teaching} have been introduced.

In an earlier work \cite{dogan_asking_2022} for HRI, the authors used an existing object detector (DETR \cite{carion2020endtoendobjectdetectiontransformers}) connected to a dense image captioning model (DenseCAP \cite{johnson2015densecapfullyconvolutionallocalization}), on which they applied GradCAM \cite{8237336}. Although this method is quite promising, it is limited by the performance of the sub-components themselves and is not specifically tailored for ambiguity detection. Furthermore, even if we wanted to finetune (fully supervised) this further using the saliency maps, we would be limited by the computational complexity of GradCAM. In fact, if we define a loss that depends on the saliency map and backpropagate, we would be computing second order gradients, which is slow and memory hungry.\\
Beyond that, recent work \cite{chisari2025robotictaskambiguityresolution} fine-tuned an end-to-end foundation model (Molmo 7B \cite{deitke2024molmopixmoopenweights}) to predict a yes/no response for ambiguity detection. However, for the balanced class ratio dataset they used, the performance was not very high (around 0.68 F1).\\
\cite{chen2025acknowledgingfocusambiguityvisual} introduced a benchmark for ambiguity detection, named “Focus Ambiguity”, which is precisely the type of ambiguity we are tackling in this paper. They showed that even large models do not achieve high performance when it comes to ambiguity classification.

\begin{figure*}[t]
    \centering
    \includegraphics[width=0.8\linewidth]{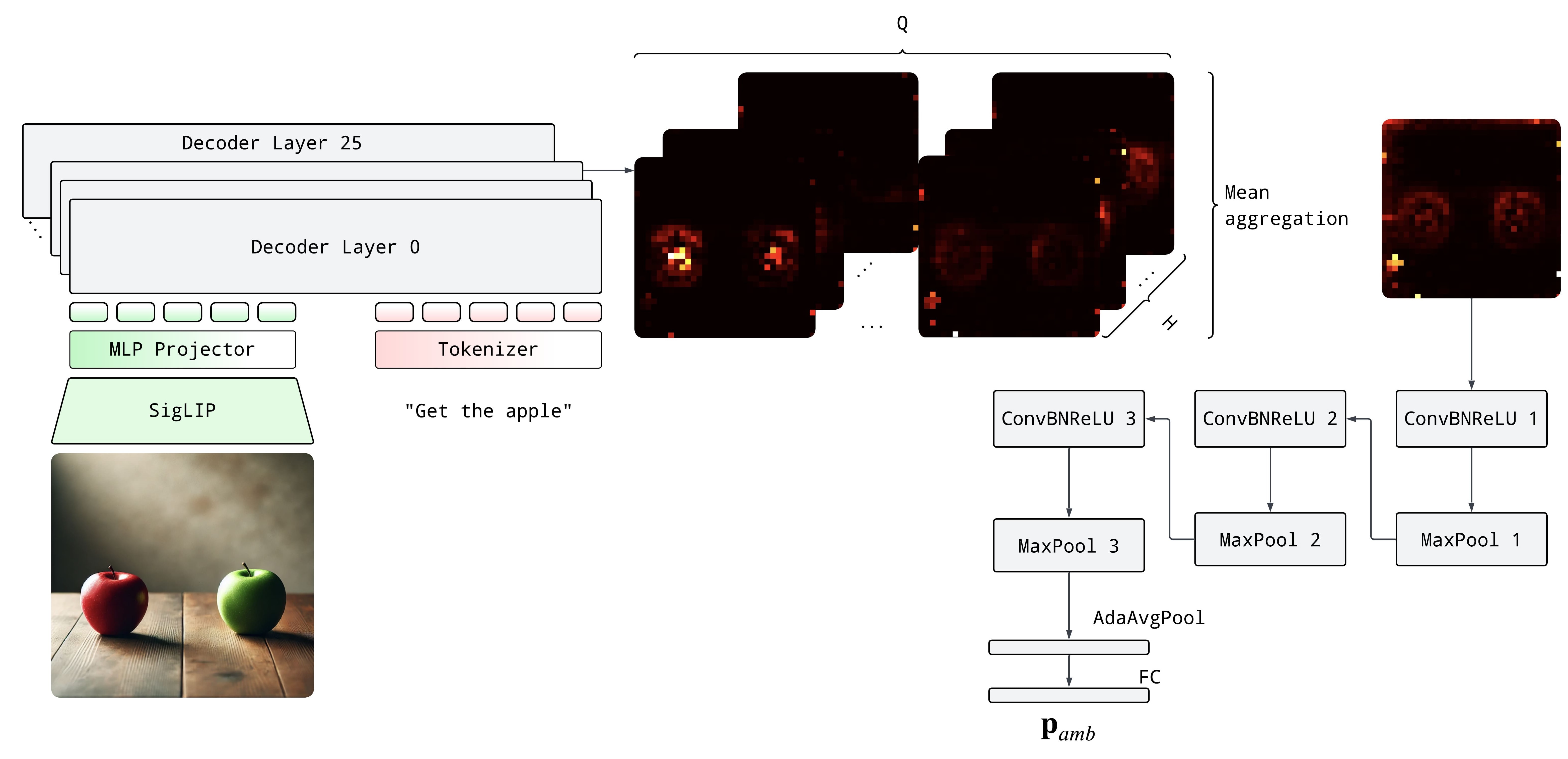}
    \caption{Ambiguity detector. The image is encoded by SigLIP and projected with an MLP. The text prefix (e.g. "Get the apple") is tokenized and passed to the Gemma2 decoder. From the 14th layer, we read text to image cross-attention for the selected prefix query tokens $Q$ and keep the first 1024 keys (image tokens), yielding per-token 32$\times$32 maps that are mean-aggregated into a single spatial map. This map is passed to a lightweight CNN to produce the ambiguity probability $p_{amb}$.} 
    \label{fig:detectarch}
\end{figure*}

\section{METHODS}
Whenever applicable to fine-tuning the VLM's language model, we use Low-Rank Adaptation (LoRA) \cite{hu2021loralowrankadaptationlarge}. Concretely, we freeze the visual encoder, the multimodal projector, and the tokenizer, and we insert rank-r LoRA adapters into the decoder's projection layers (attention q/k/v/o and MLP up/down/gate). Unless stated otherwise, we use $r=16$ and $\alpha=32$, along with a dropout rate of 0.05. We use two different adapters for each task: Adapter A for ambiguity detection and adapter B for IVG, as represented in Fig \ref{fig:generalarch}.
\subsection{CLUE architecture for IVG}
\subsubsection{Model}
We implement IVG as a transformer decoder based on PaliGemma mix 2 (paligemma2-3b-mix-448) \cite{steiner2024paligemma2familyversatile}. This model was pre-trained to handle different tasks by using a different initial token. For example, if one needs to do REC, the text instruction should be ''\textsc{\textbf{detect}} the red cup''. If we want to do image captioning, we should use ''\textsc{\textbf{caption en}} ''. Since it was also trained for open-ended object detection, its vocabulary is extended with discrete tokens $\left\{<\mathrm{lock}>\right\}_{k=0}^{1024}$ that encode normalized box coordinates in $[0,1024)$. In this line,  we reserve a special conditioning token \textsc{''\textbf{clarify}''} for our task: we attend to image tokens and the running text dialog context and autoregressively emit either a clarification question or a sequence of grounding location tokens. Figure \ref{fig:ivgarch} shows an illustration of how the model is used.\\
Let C be the running dialog context (the text prefix). Concretely, $C^{(k)}$ is the concatenation (noted $\Vert$) of the initial user request $U$ plus all turns (assistant/robot questions $R_i$ and human answers $H_i$) up to turn $k$.
At token step t, conditioned on image $I$, the linearized prefix $X$, and the previously generated suffix tokens $y_{<t}$, the decoder predicts the next token $y_{t}$.\\
The interactive exchange is a sequence of assistant/robot (model) prompts and user replies. A typical dialog is defined as:
\begin{equation*}
    D=\{(R_1,H_1), (R_2,H_2), \cdots, (R_K,H_K)\}
\end{equation*}
For training, We construct a supervised dataset $S$ of prefix $\rightarrow$ target pairs by cumulatively revealing pairs and alternating the prediction target:
\begin{align*}
  (U \rightarrow R_1), & \quad (U \Vert R_1H_1 \rightarrow R_2), \\
  & \quad \ldots, \\
  & \quad (U \Vert R_1H_1 \cdots R_{K-1}H_{K-1} \rightarrow R_K)
\end{align*}
The final grounding target would be:
\begin{equation*}
    (U\Vert R_1H_1\cdots R_KH_K\rightarrow G)
\end{equation*}
Each training example is linearized as input
\begin{equation*}
    X = \left(\texttt{"<image>clarify "}\Vert \text{prefix}\right)
\end{equation*}
and target suffix $Y$ (either a textual response $R_t$ or localization tokens $G$). We apply token-level cross-entropy only over the suffix tokens:
\begin{equation*}
    \mathcal{L}_{IVG}(\theta)=\frac{1}{T}\sum_{t=1}^T\left[-\log p_{\theta}(y_{t}|I,X,y_{< t})\right]
\end{equation*}
Where $T$ is the number of tokens in the target suffix $Y$ for this example.\\
At test time, given the initial user instruction $U$ and image $I$, we set $C_0=U$ and iterate. Algorithm \ref{alg:ivg} shows the pseudocode. 

\begin{algorithm}[t]
\caption{Interactive Visual Grounding (IVG) Inference}
\label{alg:ivg}
\begin{algorithmic}[1]
\Require Image $I$, initial user request $U$
\Ensure Bounding box decoded from a valid $\langle loc\rangle$ sequence
\State $C^{(0)} \gets U$ \Comment{running dialogue context / prefix}
\For{$k = 0,1,2,\ldots$}
    \State $X^{(k)} \gets \texttt{"<image>clarify "}\ \Vert\ C^{(k)}$
    \State $\hat{Y}^{(k)} \gets \text{Generate}(X^{(k)})$ \Comment{multiple tokens}
    \If{$\hat{Y}^{(k)}$ contains a valid $\langle loc\rangle$ sequence}
        \State \Return $\text{DecodeBox}(\hat{Y}^{(k)})$
    \Else
        \State $\hat{R}_k \gets \text{ExtractQuestion}(\hat{Y}^{(k)})$
        \State $\hat{H}_k \gets \text{GetHumanReply}(\hat{R}_k)$
        \State $C^{(k+1)} \gets C^{(k)} \,\Vert\ \texttt{"assistant: "}\Vert$
        \Statex \hspace{\algorithmicindent}$\hat{R}_k\Vert\ \texttt{"user: "}\hat{H}_k$
    \EndIf
\EndFor
\end{algorithmic}
\end{algorithm}

\subsubsection{Dataset}
We train and evaluate on the InViG-21K \cite{zhang2023invigbenchmarkinginteractivevisual} subset of the InViG benchmark for interactive visual grounding. InViG is a large-scale corpus for open-ended clarification and grounding, built in two stages: (i) 21K human-human, visually grounded dialogs collected via an online chat tool, and (ii) ~500K auto-generated human-robot disambiguation dialogs derived from the 21K seed to scale data and linguistic diversity. We use only the human-human InViG-21K subset.
\subsection{CLUE architecture for Ambiguity detection}
\subsubsection{Model}
\begin{figure*}[t]
    \centering
    \includegraphics[width=0.8\linewidth]{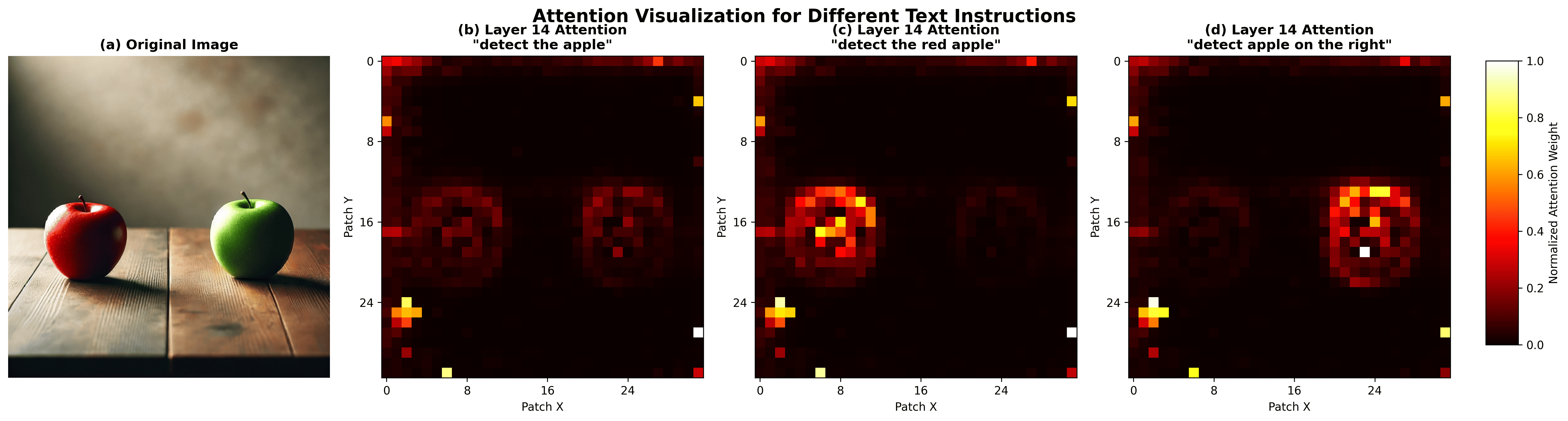}
    \caption{Text$\rightarrow$ image attention maps from layer 14 ($32\times 32$ patches) for different instructions. (a) Input image. (b) "detect the apple" yields two peaks over both apples (ambiguous). (c) "detect the red apple" concentrates on the left apple. (d) "detect apple on the right" concentrates on the right apple. Colors show min-max normalized attention weight.}
    \label{fig:attnviz}
\end{figure*}
By inspecting the pretrained detector’s (same base model) cross-modal attention, we observed that certain layers zero-shot produce sparse attention maps (Fig. \ref{fig:attnviz}). We hypothesize that this stems from the base model’s multi-object localization training: exposure to naturally ambiguous references encourages an implicit ambiguity signal that manifests in attention. On its own, the signal is insufficient for ambiguity detection because pretraining provides little explicit supervision for ambiguity. The idea is to use it as a starting point and fine-tune a model that is fully supervised on ambiguity detection. The architecture is represented in Fig. \ref{fig:detectarch}. Let the multimodal decoder backbone be a stack of $L \in \mathbb{N}$ transformer blocks $B_k$.
We define a half-depth variant $f^{1/2}$ by selecting a subset of blocks $\mathcal{S} \;=\; \{0,1,\ldots, 14\}$ (This choice is justified in Section \ref{sec:ablation}). This gives,
\begin{equation*}
    f^{(1/2)} \;=\; \mathrm{B}_{14}\circ\cdots\circ\mathrm{B}_{1}\circ\mathrm{B}_{0}
\end{equation*}
Given an image $I$ and instruction $x$, we run a forward pass $f^{1/2}$ and extract the attention tensor at the chosen block: $A\in \mathbb{R}^{H\times Q\times K}$, where $H$ is the number of heads in the Grouped Query self-attention \cite{vaswani2023attentionneed, ainslie2023gqatraininggeneralizedmultiquery}, Q is the query length (text tokens), and K is the key length (image+text tokens). We create a mask over query positions $\mathcal{Q}_x=\{q\in {1, \ldots,Q}: token(q)\notin\mathcal{E}\}$, excluding special tokens $\mathcal{E}$ (conditioning token, eos, pad). 
To prevent heads with high raw attention magnitudes from dominating the map, we perform per-head L1 renormalization over the image attention. Let $L_{img}$ be the number of image tokens (patches). For each head $h$ and query $q \in \mathcal{Q}_x$, we restrict keys to the image region $k \in \{1, \dots, L_{img}\}$ and normalize:
\begin{equation}
    \bar{A}_{h,q,k} = \frac{A_{h,q,k}}{\sum_{j=1}^{L_{img}} A_{h,q,j} + \epsilon}
\end{equation}
This ensures that the attention distribution sums to 1 over the image for every head. We then mean-aggregate these normalized maps:
\begin{equation}
    v = \frac{1}{|\mathcal{Q}_x|H}\sum_{h=1}^H\sum_{q\in \mathcal{Q}_x}\bar{A}_{h,q,:}\in \mathbb{R}^{L_{img}}
\end{equation}
We use mean aggregation for query-token fusion because PaliGemma applies bidirectional self-attention over the prefix (instruction) and a causal mask only on the suffix (See Fig. 2 in \cite{beyer2024paligemmaversatile3bvlm}). Thus, the instruction tokens (e.g., apple, left) fully attend to one another, yielding contextualized token embeddings by the time we read cross modal attention to image patches. Averaging these queries post Softmax is therefore a simple, stable and differentiable way to capture the composed semantics that already emerge from their mutual interaction.\\
Let $v_{img}\in \mathbb{R}^K_{img}$ be the prefix of v corresponding to images keys (patch tokens). We reshape to a spatial grid, $32\times 32$ in our case.
After per-sample min-max normalization,
a lightweight probe $g_\theta:[0,1]^{H_M\times W_M}\rightarrow[0,1]$ maps the attention map $M$ to the probability of ambiguity $p_{amb}$ (See Fig. \ref{fig:detectarch}). 
Given labels $y\in\{0,1\}$ (1=ambiguous, 0=unambiguous), we minimize binary cross-entropy over a batch $\mathcal{B}$:
\begin{align*}
\mathcal{L}_{amb}(\theta, \phi, \psi) 
&= \frac{1}{|\mathcal{B}|} \sum_{(I,x,y)\in\mathcal{B}} 
   \Big[ -y \log p_{amb} \nonumber \\
&\qquad\quad - (1-y)\log(1-p_{amb}) \Big]
\end{align*}
Here $\phi$ are frozen backbone weights and $\psi$ are LoRA adapter parameters. 
\subsubsection{Dataset}
\label{sec:dataset}
For this task, we generated synthetic data using the Isaac Sim simulator \cite{NVIDIA_Isaac_Sim}, consisting of 2000 tabletop scenes populated with randomized YCB objects. To construct the corresponding textual descriptions, we enforced a constraint during randomization such that each scene contains at least two visually similar objects. In this setting, ambiguous instructions simply take the form \textsc{“Get the $\prec$duplicated$\_$object$\succ$”}, whereas unambiguous instructions correspond to expressions such as \textsc{''Get the $\prec$unique$\_$object$\succ$''}. Moreover, we combined it with another dataset (IT2P) from \cite{ahn_interactive_2018}, which has 477 images, totaling 1847 image-text samples. In total, the training dataset (Dataset 1) includes around 4000 image–instruction pairs. We further set aside 100 manually labeled real-world images from InVIG \cite{zhang2023invigbenchmarkinginteractivevisual} for evaluation. Each image has both an ambiguous and an unambiguous label, giving a total of 200 image-instruction samples (Dataset 2). The models were not trained on this dataset and we consider it Out-of-distribution (OOD).
\begin{figure}
    \centering
    \includegraphics[width=1\linewidth]{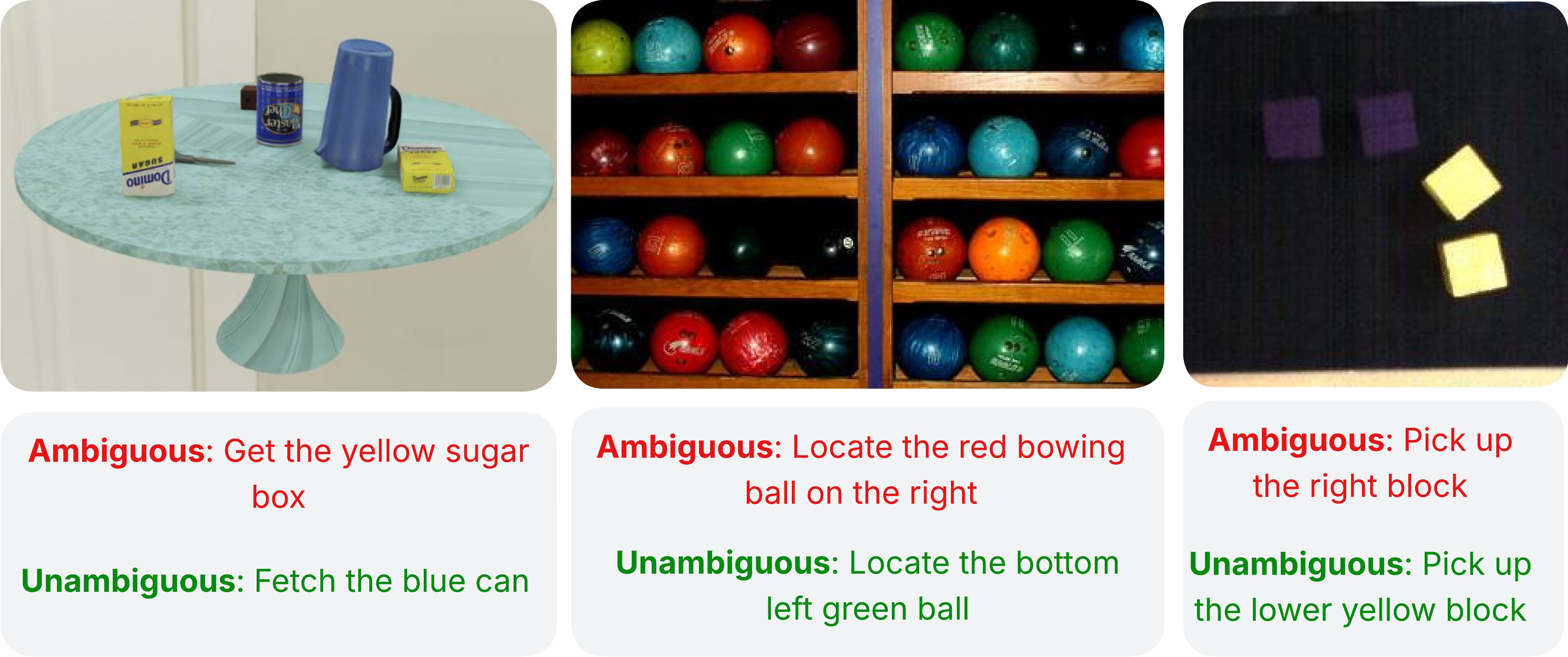}
    \caption[Samples from the ambiguity detection dataset.]{Samples from the ambiguity detection datasets. (Left) Synthetic data from Isaac Sim, (Middle) real-world data (Dataset 2 (OOD), (Right) IT2P: dataset from \cite{ahn_interactive_2018}. Dataset 1 is a combination of samples from (Left) and (Right)}
    \label{fig:dataset}
\end{figure}
\section{RESULTS AND DISCUSSION}
\subsection{Ambiguity detection}
\subsubsection{Ablation study}
\label{sec:ablation}
\begin{figure}
    \centering
    \includegraphics[width=\linewidth]{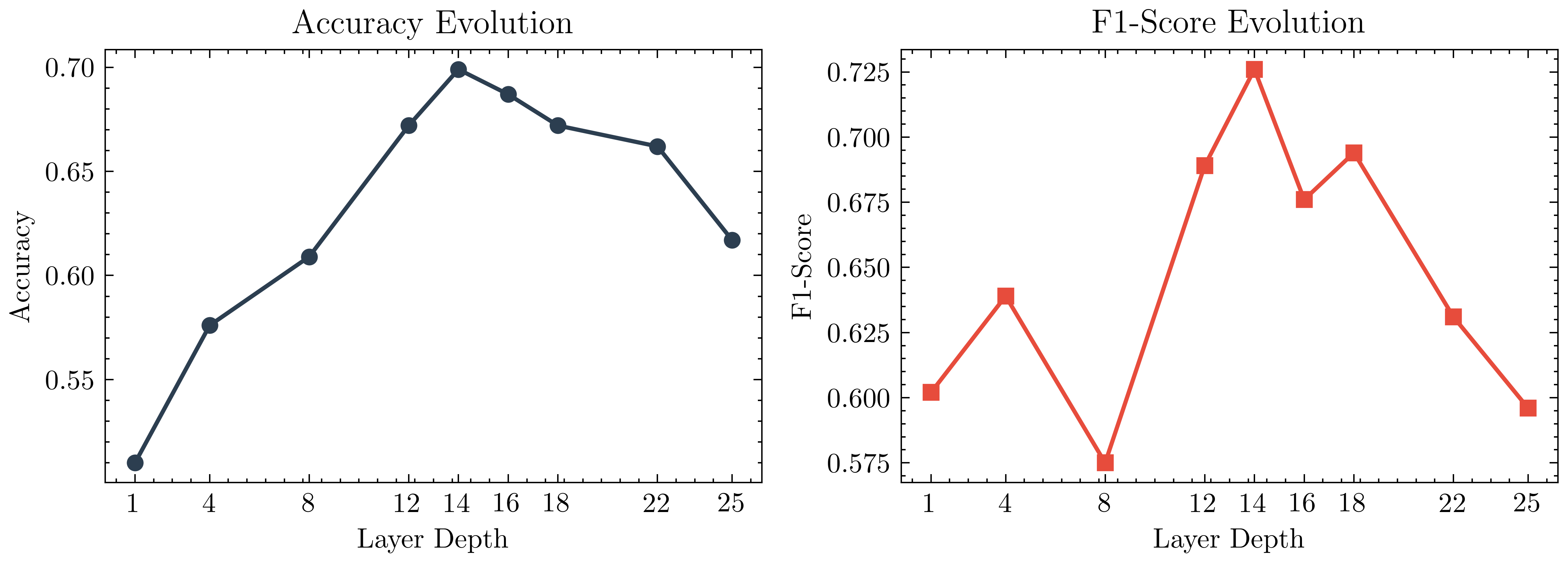}
    \caption{F1-score and accuracy as a function of the number of the decoder layer used}
    \label{fig:f1vslayer}
\end{figure}

To understand how the choice of the decoder layer affects the attention signal and to justify the choice of layer 14, we freeze all parts of the model except the CNN head and then train the model while changing the decoder layer used to extract that signal. Figure \ref{fig:f1vslayer} shows the F1 score across the gradually increasing number of decoder layers. As can be seen, the F1 score improves steadily from the early stages, peaking significantly around Layer 14 (0.726), where the model achieves its optimal balance of Precision and Recall. However, a performance decrease is observed at the final stage, Layer 25, where the F1 score drops to 0.596.
\subsubsection{Setup}
\begin{table*}[t]
\centering
\caption{Performance comparison of different models on Dataset 1 and Dataset 2.}
\label{tab:results}
\renewcommand{\arraystretch}{1.2}
\small
\setlength{\tabcolsep}{12pt}
\begin{tabular}{lcccc|cccc}
\toprule
\multirow{2}{*}{\textbf{Model}} & \multicolumn{4}{c|}{\textbf{Dataset 1 (Test)}} & \multicolumn{4}{c}{\textbf{Dataset 2 (OOD)}} \\
\cmidrule(r){2-5} \cmidrule(l){6-9}
 & \textbf{Acc} & \textbf{Prec} & \textbf{Rec} & \textbf{F1} & \textbf{Acc} & \textbf{Prec} & \textbf{Rec} & \textbf{F1} \\
\midrule
\rowcolor{lightgray} \multicolumn{9}{l}{\textit{Detect Instruction (Half-Depth)}} \\
Half-Last Detect (CNN) & 0.850 & 0.847 & 0.846 & 0.846 & 0.732 & 0.680 & 0.874 & 0.765 \\
Half-Last Detect (AR) & 0.729 & 0.795 & 0.599 & 0.683 & 0.674 & 0.636 & 0.811 & 0.713 \\
\midrule
\rowcolor{lightgray} \multicolumn{9}{l}{\textit{Disambiguate Instruction (Half-Depth)}} \\
Half-Last Disambig. (AR) & 0.874 & 0.839 & 0.917 & 0.876 & 0.763 & 0.695 & \textbf{0.937} & 0.798 \\
Half-Full Disambig. (AR) & \textbf{0.912} & \textbf{0.934} & 0.881 & \textbf{0.907} & \textbf{0.826} & \textbf{0.793} & 0.884 & \textbf{0.836} \\
\midrule
\rowcolor{lightgray} \multicolumn{9}{l}{\textit{Disambiguate Instruction (Full-Depth)}} \\
Full-Last Disambig. (AR) & 0.701 & 0.629 & 0.940 & 0.754 & 0.632 & 0.634 & 0.621 & 0.628 \\
Full-Full Disambig. (AR) & 0.893 & 0.893 & 0.887 & 0.890 & 0.647 & 0.608 & 0.832 & 0.702 \\
\midrule
\rowcolor{lightgray} \multicolumn{9}{l}{\textit{Zero-Shot Baselines}} \\
Gemma3-4B-it & 0.487 & 0.487 & \textbf{1.000} & 0.655 & 0.500 & 0.500 & \textbf{1.000} & 0.667 \\
Gemma3-27B-it & 0.609 & 0.560 & 0.931 & 0.699 & 0.479 & 0.486 & 0.747 & 0.589 \\
\bottomrule
\end{tabular}%

\end{table*}
We evaluated the ambiguity detector on the synthetic dataset of instruction-image pairs, in which the scenes contain visually similar distractors. An example is labeled ambiguous if the instruction under-specifies a unique referent; otherwise, it is unambiguous. We use a 70/15/15 split and identical preprocessing for all methods. We used the AdamW optimizer with a learning rate of 5e-6 for the LoRA adapters and 1e-4 for the CNN head. We applied a weight decay of 1e-4. Training was done on a single A100 GPU.
\subsubsection{Baselines}
Because the pretraining for REC was conditioned on the token “detect”, using it gave the most interpretable signal (see Fig. \ref{fig:attnviz}). We therefore used “detect” for the CNN-based model and some autoregressive variants (where the decoder has to autoregressively generate "yes" or "no"), and introduced a new token, “disambiguate”, unused during pretraining, for the remaining variants.
In detail, we performed a comparative analysis of several model variants. To ensure clarity, we adopt a naming convention denoting \textbf{Model depth} (Half vs. Full decoder layers), \textbf{Finetuning strategy} (Last layer vs. Full network), and \textbf{Instruction type}.
\begin{itemize}
    \item \textbf{Half-Last Detect (CNN)}: This follows the model in Fig. \ref{fig:detectarch}. A CNN classifier trained on attention maps extracted from around the first half of the decoder layers ("Half"), and only the last decoder layer is finetuned along with the CNN. It uses the instruction starting with "detect" (seen during pretraining).
    
    \item \textbf{Half-Last Detect (AR)}: An autoregressive approach using only the first half of the decoder layers. Only the last layer is finetuned, using the standard "detect" instruction.
    
    \item \textbf{Half-Last Disambig. (AR)}: Uses the first half of the decoder layers with only the last layer finetuned. However, it uses a new instruction starting with "disambiguate" (unseen during pretraining) to test generalization.
    
    \item \textbf{Half-Full Disambig. (AR)}: Uses the first half of the decoder layers, but unlike the "Last" variants, all parameters in these layers are LoRA finetuned using the "disambiguate" instruction.
    
    \item \textbf{Full-Last Disambig. (AR)}: Uses the complete set of decoder layers ("Full" depth). Only the very last layer is finetuned ("Last") using the "disambiguate" instruction.
    
    \item \textbf{Full-Full Disambig. (AR)}: The full decoder depth with all layers LoRA finetuned ("Full" finetuning) with the "disambiguate" instruction.
    
    \item \textbf{Gemma-it Baselines}: We include zero-shot evaluations of Gemma3-4B-it and Gemma3-27B-it \cite{gemmateam2025gemma3technicalreport} to provide a reference for standard instruction-tuned performance on this task.
\end{itemize}
The goal of this comparison is to see if using the internal representation of VLM provides a useful signal for ambiguity detection and up to which level. 
\subsubsection{Results}
In the finetuned setting, we observe that under the same architecture and “Detect” instruction strategy, the CNN-based model significantly outperforms the classic Auto-Regressive (AR) baseline. As shown in Table \ref{tab:results}, the Half-Last Detect (CNN) achieves an F1-score of 0.846, compared to just 0.683 for the autoregressive approach. Furthermore, our results confirm that using around half the layers is actually superior to using the full layers; for instance, the AR “Disambiguate” task drops from 0.876 F1 at half-depth (Half-Last) to 0.754 at full-depth (Full-Last). 
Moreover, although \textbf{Full-Full Disambig. (AR)} attains almost the same in-domain performance as \textbf{Half-Full Disambig. (AR)} (0.89 vs. 0.907), its score drops sharply to 0.702 on the real-world OOD dataset, compared to 0.836 at half layers, indicating that middle layers exhibit greater robustness to domain shift than later layers.

\begin{figure}
    \centering
    \includegraphics[width=1\linewidth]{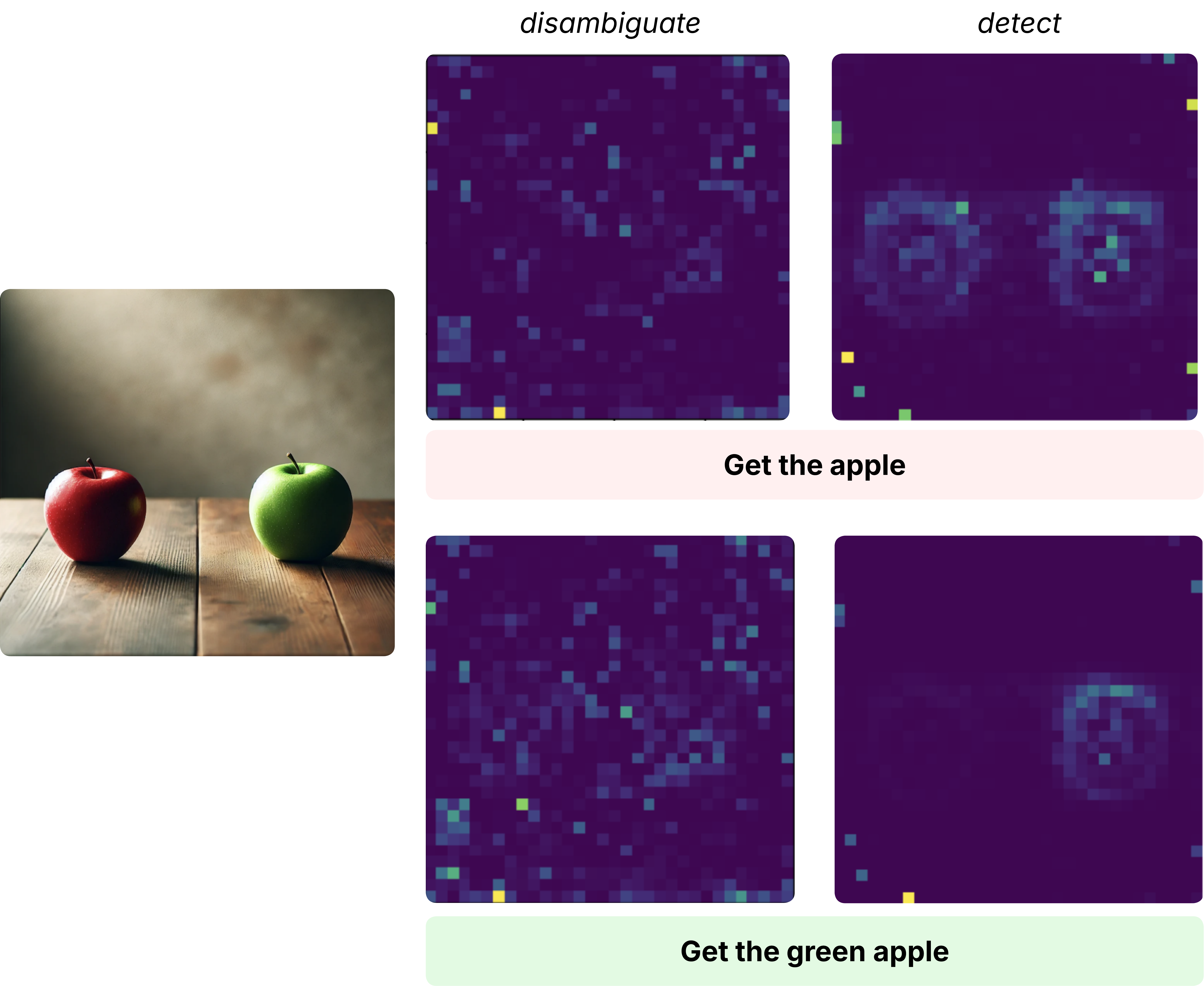}
    \caption[Visualization of middle attention maps for the "disambiguate" and "detect" tokens]{Visualization of middle attention maps for the "disambiguate" and "detect" tokens. It can be seen that the former yields noisy and unexplainable signals, while the "detect" token gives meaningful, explainable signals.}
    \label{fig:xpl}
\end{figure}

\begin{figure}
    \centering
    \includegraphics[width=\linewidth]{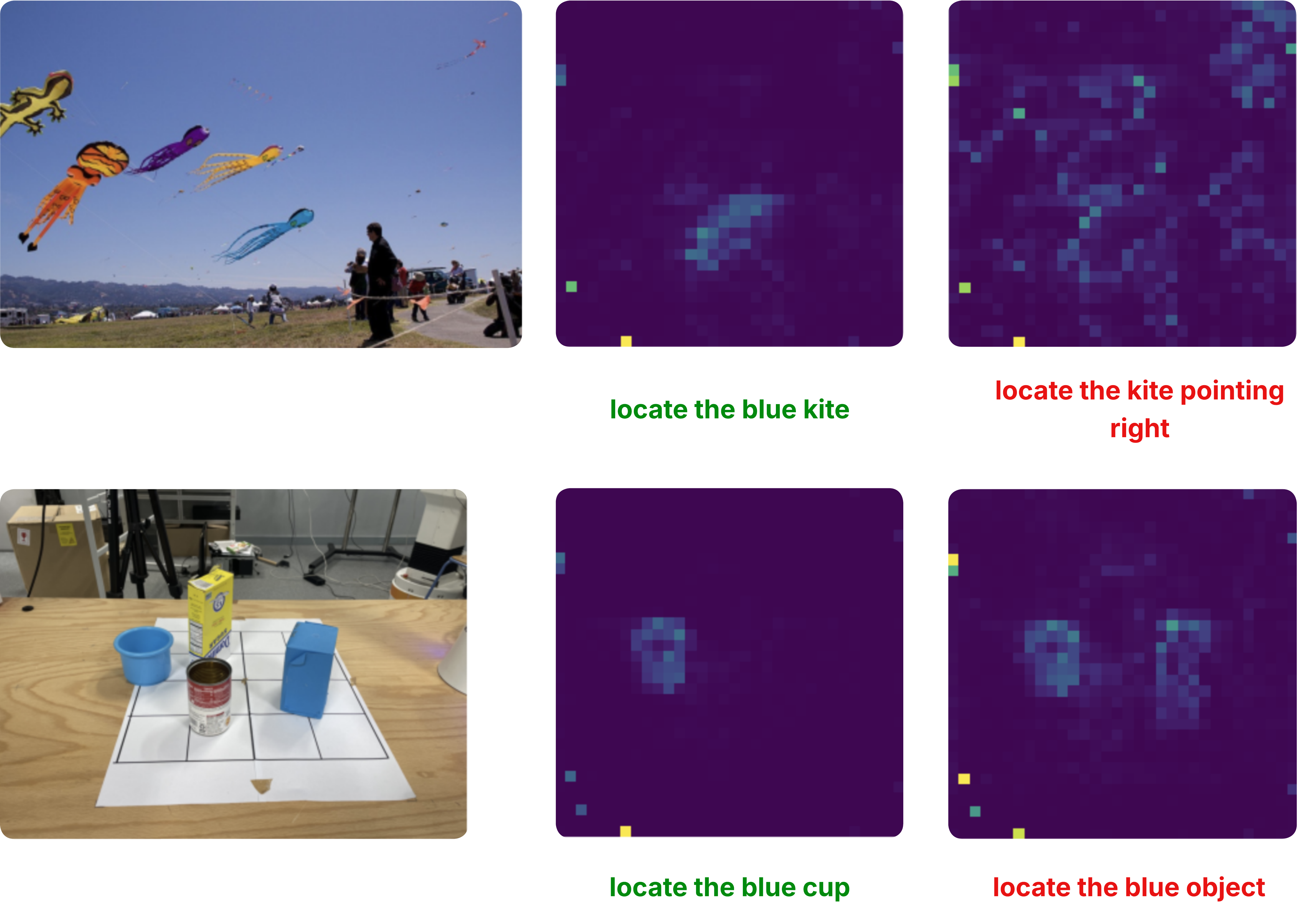}
    \caption[Qualitative examples of the attention map ambiguity signal after training from the Half-Last Detect (CNN) model]{Qualitative examples of the attention map ambiguity signal after training for the Half-Last Detect (CNN) model.}
    \label{fig:qualdetect}
\end{figure}

Furthermore, while the generative “Disambiguate” task at half-layers slightly outperforms the \textbf{Half-Last Detect (CNN)} detection (0.876 vs. 0.846), this comes with a cost. Since the “disambiguate” conditioning token was never seen during pretraining, with finetuning, the model shapes its attention to achieve the task without producing a meaningful, explainable signal. This is illustrated in Figure \ref{fig:xpl}.
Figure. \ref{fig:qualdetect} shows some qualitative results of the model.
\subsubsection{Generalization}
We observe that overall performance slightly drops on Dataset 2 (real-world OOD); for example, the \textbf{Half-Last Detect (CNN)} drops from 0.846 to 0.765. However, the results are consistent and correlated with the performance on Dataset 1, suggesting that the learned ambiguity signals remain robust even if absolute accuracy is impacted by domain shifts. In comparison, zero-shot baselines remain limited, with Gemma-27B scoring only 0.699 on Dataset 1 and 0.589 on Dataset 2, significantly lagging behind our fine-tuned approach.
\subsubsection{Comparison with SOTA}
The dataset used in this section differs from the ones used in \ref{sec:dataset} and is drawn from the work in \cite{chisari2025robotictaskambiguityresolution}, which includes both a synthetic dataset and a real-world dataset.
As can be seen in Table \ref{tab:compsota}, our model consistently outperforms the AmbResVLM \cite{chisari2025robotictaskambiguityresolution} baseline in both simulation and real-world settings. While near-perfect performance is achieved in simulation, the model also demonstrates good performance on the real world dataset. We attribute the significant gap between our model and \cite{chisari2025robotictaskambiguityresolution} to the fact that the size of the VLM used in their case (Molmo 7B \cite{deitke2024molmopixmoopenweights}) is more prone to overfitting if hyperparameters are not carefully chosen. 
\begin{table*}
\centering
\caption{Performance comparison between our model and AmbResVLM \cite{chisari2025robotictaskambiguityresolution}.}
\label{tab:compsota}
\renewcommand{\arraystretch}{1.3}
\small
\begin{tabular}{lcccccccc}
\hline
\multirow{2}{*}{\textbf{Model}} 
& \multicolumn{4}{c}{\textbf{Simulation}} 
& \multicolumn{4}{c}{\textbf{Real World}} \\
\cline{2-9}
& \textbf{Acc} & \textbf{Prec} & \textbf{Rec} & \textbf{F1}
& \textbf{Acc} & \textbf{Prec} & \textbf{Rec} & \textbf{F1} \\
\hline
AmbResVLM \cite{chisari2025robotictaskambiguityresolution}
& -- & 0.52 & 0.97 & 0.68
& -- & 0.48 & 0.78 & 0.59 \\
\hline
\textbf{Half-Full Disambig. (AR)}
& \textbf{0.995} & \textbf{1.000} & \textbf{0.989} & \textbf{0.994}
& \textbf{0.795} & \textbf{0.898} & \textbf{0.644} & \textbf{0.750} \\
\hline
\end{tabular}
\end{table*}
\subsection{Interactive Visual Grounding}
\subsubsection{Setup}
Following \cite{xu2024unifiedinteractivevisualgrounding}, we evaluate in the guesser setting: the model receives the full dialog history (instruction + all clarifications and answers) and must output a bounding box. A prediction counts as correct if Intersection over Union (IoU) $\geq$ 0.5 with the ground truth. We report Acc@0.5. Decoding uses our location-token vocabulary, and greedy decoding is applied for bounding box tokens (questions are irrelevant here since the dialog history is given).
We fine-tuned the same pretrained VLM in its mix and standard non-mix version (mix was trained on REC also) on InViG-21K only, varying LoRA capacity while freezing the vision tower and projector. We used a learning rate of 1e-4 with a weight decay of 1e-4 and a linear LR scheduler with a warmup phase. Fine-tuning was done on 8 A100 GPUs and took around 2 hours. We used the same train/val split for all the runs.
\subsubsection{Baseline}
For this comparison, we evaluate against TiO \cite{xu2024unifiedinteractivevisualgrounding} in the configuration trained from scratch on InViG only. TiO uses a transformer encoder-decoder architecture. Our goal is to test whether parameter-efficient fine-tuning (PEFT/LoRA) of a pre-trained VLM can match or even surpass a model trained from scratch under the same data budget. Fine-tuning reuses broad visual-linguistic priors, typically converges faster, requires far fewer trainable parameters/compute, and is easier to adapt across tasks or domains. In low-to-moderate data regimes, this can be preferable to building and training an architecture from scratch.
\subsubsection{Results}
\begin{figure}
    \centering
    \includegraphics[width=0.8\linewidth]{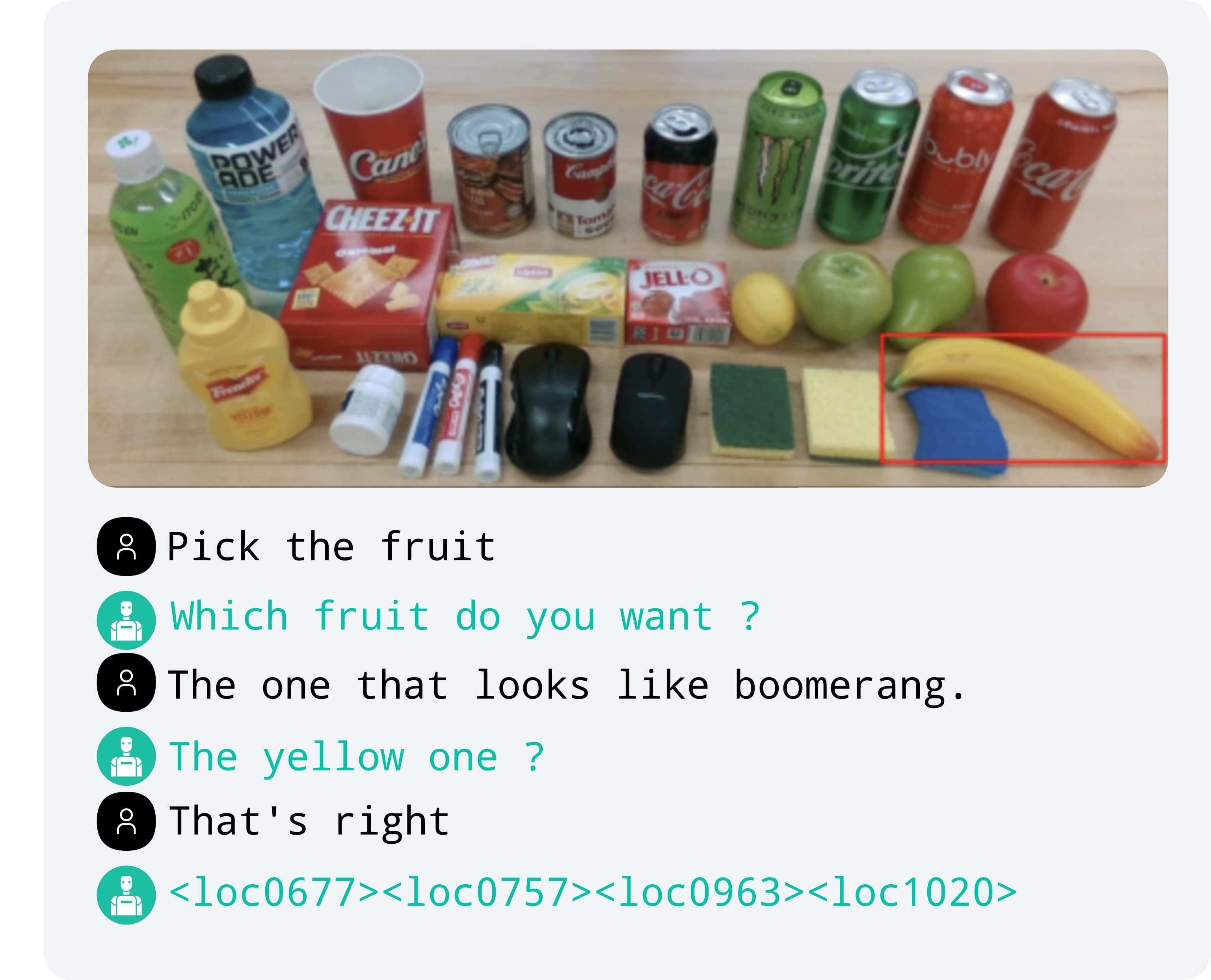}
    \caption{Qualitative result of IVG.}
    \label{fig:qualivg}
\end{figure}

We evaluate our LoRA fine-tuned CLUE model against the TiO baseline, which scores 71.2$\%$ Acc@0.5 on InViG-only training (Table \ref{tab:table2}). Our results highlight the critical role of pre-training mixtures. The CLUE variant fine-tuned from the general PaliGemma-2 checkpoint (non-mix) underperforms the baseline at 63.19$\%$. However, variants fine-tuned from checkpoints that include object detection data (mix) achieve significantly higher score of 75.66$\%$. Our best model demonstrates an improvement of 4.46$\%$ over the TiO baseline.
This large gap between the non-mix and mix variants strongly supports the hypothesis that object detection training provides essential spatial priors that boost IVG performance \cite{xu2024unifiedinteractivevisualgrounding}. Furthermore, we observe that performance is robust to adapter capacity in our mix settings. Ultimately, we demonstrate that parameter-efficient fine-tuning of an appropriately pre-trained VLM can be a more effective and compute-efficient strategy than training from scratch for IVG tasks, especially in low-data regimes.
Fig. \ref{fig:qualivg} shows a qualitative result for IVG.
\begin{table}[]
\centering
\begin{tabular}{llll}
\hline
Model                          & LoRA config  & Acc@0.5 \\ \hline
\textbf{TiO \cite{xu2024unifiedinteractivevisualgrounding}} &     None      &     71.2$\%$     \\ \hline
\textbf{CLUE (non-mix backbone)} & $\alpha=8$, $r=16$         &     63.19$\%$     \\ \hline
\textbf{CLUE (mix)} & $\alpha=32$, $r=16$         &     75.55$\%$     \\ \hline
\textbf{CLUE (mix)}     &   $\alpha=16$, $r=8$      &     \textbf{75.66$\%$  }   \\ \hline
\end{tabular}
\caption{Comparison between IVG models trained on InViG only.}
\label{tab:table2}
\end{table}

\section{CONCLUSION AND FUTUR WORK}
We introduced an ambiguity-aware IVG pipeline that (i) reads cross-modal attention over image patches from a PaliGemma2 backbone and trains a lightweight CNN to detect ambiguous instructions, and (ii) fine-tunes a pretrained VLM with LoRA to perform clarification dialogue and output grounding location tokens. On InViG-only training, CLUE surpasses a state-of-the-art baseline trained from scratch, while the attention-map detector achieves strong accuracy with half-depth compute and outperforms baselines.
Future work could  include joint training by merging the ambiguity head with the IVG decoder using an auxiliary loss, so that clarification behaviour and detection can co-adapt.
All in all, we see attention-based ambiguity signals as an, interpretable bridge between black-box VLMs and HRI: they are relatively cheap to compute, naturally grounded in the scene.

\section*{Acknowledgments}
This research was partially funded by the French National Research Agency (ANR) under the OSTENSIVE (ANR-24-CE33-6907-01) and ANITA (ANR-22-CE38-0012-01) projects

\addtolength{\textheight}{-12cm}   

\bibliographystyle{IEEEtran}
\bibliography{IEEEabrv}

\end{document}